# Evaluative Judgement in Teaching AI-based Translation: A Classroom Case Study of AI-Mediated Translation and Post-Editing

**Gokhan Dogru**[1]
Universitat Pompeu Fabra
Barcelona, Spain
gokhan.dogru@upf.edu

## Abstract

Drawing on 23 anonymized student projects from a fourth-year Machine Translation and Post-editing course in a BA-level translation programme, this paper examines how structured comparison of general-purpose LLMs and online MT systems can elicit evaluative judgement in AI-mediated translation. Students translated short specialised English Wikipedia texts into Catalan or Spanish, generated four system outputs, evaluated them using automatic metrics and human adequacy/fluency assessment, selected one output for post-editing, and justified their decision in written reports. Descriptive counts are reported for all 23 projects, while qualitative interpretation is based on the 22 cases accompanied by written reports. Results show that students did not treat automatic metrics as final authority: final post-editing selections often diverged from metric rankings and were justified through adequacy, fluency, terminology, naturalness, and expected post-editing effort. The study therefore does not benchmark systems under controlled conditions; it analyses how students justified system choice within an authentic classroom assignment.

## 1. Introduction

Translation technology pedagogy has long argued for evaluation, comparison, and post-editing as central elements of translator training, alongside tool operation. Earlier work in this field had already argued that students should learn not only how to operate translation tools and machine translation (MT) systems, but also how to evaluate them critically, compare technologies, and understand post-editing as a professional and judgement-based activity rather than a merely mechanical one (Ayvazyan et al., 2024). More recent studies build on this perspective by emphasizing that translator education must now engage explicitly with AI literacy, including issues such as human agency, prompting practices, data awareness, trust, and ethical decision-making (Freeth and Toto, 2026; Zhang and Doherty, 2025; European Commission, 2022; Massey and Ehrensberger-Dow, 2026).

The case study reported here comes from a fourth-year university course on Machine Translation and Post-editing. The aim is not to determine which system performs best in general terms, but to examine how a structured assignment can support students in developing "evaluative judgement" (Tai et al., 2018) in AI-mediated translation. The assignment had five stages. Students first produced a human translation, then generated four system outputs, evaluated them with automatic metrics and human adequacy/fluency criteria, selected one version for post-editing, and justified that decision in a report. This design follows recent calls for classroom experimentation in response to technological change, particularly tasks in which students compare human, machine-generated, and post-edited versions rather than treating AI/MT output as a finished product (Ayvazyan et al., 2024; Kenny, 2020; Guerberof Arenas and Moorkens, 2019).

The problem is especially relevant in a context where AI-generated translations may appear fluent, coherent, and convincing while still containing terminological, adequacy-related, or stylistic problems that demand careful human scrutiny (Ayvazyan, 2025). In line with recent higher education research, this study treats evaluative judgement as a core graduate capability: the ability to make informed decisions about the quality of work and to justify those decisions using explicit criteria and evidence. In the context of GenAI, this becomes even more important, since students must remain the "arbiters of quality" (Bearman et al., 2024, p. 894) rather than uncritical accepters of machine output. The assignment is valuable because students had to leave a trace of their reasoning: metric interpretation, error comments, system selection, and post-editing choices.

The study therefore contributes an integrated classroom design for teaching critical AI use in translator education. More specifically, it examines how students compare translation outputs, interpret metric scores in relation to human judgement, identify recurring error patterns, and justify why one version is more suitable than others for post-editing. In doing so, the paper argues that structured comparison and evidence-based selection can serve as an effective pedagogical means of fostering critical, reflective, and professionally relevant judgement in the age of GenAI.

This study is guided by three research questions:

RQ1. How did students justify their final selection of one LLM/MT output for post-editing?

RQ2. To what extent did students' final post-editing choices converge with or diverge from automatic metric rankings?

RQ3. What kinds of evaluative judgement do students articulate when comparing LLM and NMT outputs within a translation/post-editing assignment?

The unit of analysis is therefore not the translation system as such, but the student evaluative decision: how students interpreted different forms of evidence and justified the choice of one output as a post-editing candidate.

## 2. Related Work

### 2.1. From MT literacy to AI literacy in translator education

Translation technology pedagogy has increasingly moved from procedural tool instruction toward literacy-oriented models of translator education including MT literacy, AI literacy and data literacy (Krüger and Hackenbuchner, 2024; Kenny, 2020; Rico and González Pastor, 2022; Moorkens et al., 2024). In these models, competence includes understanding how systems work, when they are appropriate, and what consequences follow from using them in specific communicative and professional contexts. O'Brien and Ehrensberger-Dow (2020), for example, conceptualize MT literacy as a contextualized and cognitively grounded ability to understand what MT does, what it does not do, and how its use affects translation processes and decisions. Ehrensberger-Dow et al. (2023) extend this perspective by proposing the role of translators and trainers as "MT literacy consultants," arguing that education should mitigate the risks of uninformed MT use by cultivating informed judgement and guidance.

Recent scholarship on GenAI broadens this literacy agenda further. Penet et al. (2026) argue that GenAI is not merely another translation tool, but a development that requires translator education to re-engage questions of human agency, ethics, and sustainability. Within that volume, AI literacy is framed in explicitly translation-specific terms through the notions of suitability, data literacy, prompting, and ethical reasoning (Bowker, 2026; Inglada, 2026; Moorkens and Doğru, 2026; Yamada, 2026). This translation-oriented work aligns with broader AI literacy research, where AI literacy is defined as a set of competencies that enable people to critically evaluate AI systems and interact with them responsibly (Long and Magerko, 2020). It also resonates with recent empirical evidence showing that novice translation students already use AI tools early in their training, often before they have developed robust critical or ethical routines for assessing output quality (Zhang and Doherty, 2025). As a whole, these studies suggest that translator education should not simply teach students how to obtain AI-generated output, but how to judge when, how, and why such output should be used, compared, and revised in context (Ayvazyan et al., 2024).

### 2.2. Evaluation and post-editing as pedagogical practices

The pedagogical value of evaluation and post-editing has a longer history in translator education. Early EAMT work already treated both MT evaluation and post-editing as legitimate objects of instruction and assessment (Belam, 2002; O'Brien, 2002). Later studies further embedded MT and post-editing into translation curricula through explicit learning objectives, professional workflows, and classroom tasks designed to connect academic training to industry practice (Doherty and Kenny, 2014; Guerberof Arenas and Moorkens, 2019). Seen in this wider context, evaluation and post-editing pedagogy are not merely ways of familiarising students with professional workflows; they also cultivate the added value of human translators as adaptive and digitally literate experts whose role increasingly lies in reading, prompting, evaluating, editing, and making informed decisions about how AI-supported output should be used (Massey et al., 2023; Massey and Ehrensberger-Dow, 2026).

Within this tradition, evaluation is especially important because it shifts students from passive tool users to active judges of quality. Moorkens (2018) proposes a practical in-class NMT evaluation exercise that combines adequacy, post-editing productivity, and a simple error taxonomy in order to demystify technological hype and foster critical judgement. Öner and Öner Bulut (2021) similarly report that a post-editing-oriented NMT quality-evaluation and error-annotation task can support translator training by prompting students to identify, categorize, and reflect on translation problems more systematically, rather than relying only on general impressions of whether a translation "sounds good." At the same time, the post-editing literature warns against assuming that such practices are self-explanatory to novices. Flanagan and Christensen (2014) show that trainees do not always interpret post-editing guidelines consistently, which means that judgement criteria and revision expectations must be explicitly scaffolded. Nitzke et al. (2019) add a further dimension by framing post-editing competence as risk management: competent translators must judge not only how to post-edit, but also whether MT output is suitable for a given task, text type, and quality requirement. This perspective is particularly relevant for pedagogical designs in which students must compare several outputs and justify the selection of one version for post-editing.

## 2.3. Automatic metrics, human judgement, and evaluative judgement

A second core strand of related work concerns how quality is judged and how students learn to interpret different forms of evidence. In higher education research, evaluative judgement is understood as the capacity to make decisions about the quality of one's own work and that of others (Tai et al., 2018). In the context of GenAI, this capacity becomes even more important because fluent and persuasive outputs can mask deeper adequacy, factual, or domain-specific problems (Bearman et al., 2024). This insight is highly transferable to translation pedagogy, where students must learn to distinguish surface fluency from fuller notions of translation-oriented quality.

Automatic metrics are relevant here not because they replace human judgement, but because they provide one source of evidence that students must interpret critically. Rei et al. (2020) introduce COMET as a learned evaluation framework designed to correlate more strongly with human judgements than older lexical-overlap metrics. However, meta-evaluation work from WMT23 shows that metric–human correlation remains sensitive to evaluation conditions, including the quality of the reference translation itself (Freitag et al., 2023). For classroom practice, the implication is concrete: students need to learn why a high COMET, BLEU, or ChrF score may still coexist with terminology, adequacy, or naturalness problems. In this respect, structured error annotation provides an important complement to both metrics and impressionistic human evaluation. MQM offers an established framework for making error categories explicit and discussable, even when simplified for educational purposes (Lommel et al., 2014). When used together, metrics, adequacy/fluency ratings, and error analysis give students several kinds of evidence to reconcile.

## 2.4. Ethics, trust, and agency

Ethics is also part of this pedagogical problem. Ramírez-Polo and Vargas-Sierra (2023) show that ethical competence is often under-explicit in translation technology syllabi, despite the growing relevance of issues such as confidentiality, bias, accountability, and responsible use. Tipton (2024) similarly argues for integrating ethics into classroom tasks and reflective practice rather than treating it as detached policy knowledge. In translator education specifically, Moorkens and Doğru (2026) frame AI ethics as a practical discipline of deciding when GenAI should and should not be used, linking ethics directly to classroom reasoning and decision-making. This position is consistent with broader educational guidance that advocates human-centred and risk-aware approaches to GenAI in learning environments (Miao and Holmes, 2023).

The ethics dimension also intersects with questions of trust and professional agency. Rivas Ginel and Moorkens (2025) show that translators' trust and distrust in GenAI are shaped not only by perceived quality, but also by accountability, risk, and appropriate use. For translator education, this suggests that students should not merely be encouraged to use AI tools, but should be trained to calibrate reliance on them and to justify that reliance explicitly. Within this framework, the pedagogical task is not to identify a universally best system, but to help students make accountable decisions about quality, intervention, and responsibility while retaining a sense of human control and agency in their interaction with AI tools.

Existing work has established the pedagogical value of MT evaluation, post-editing, and AI literacy in translator education. However, there is still limited empirical evidence on classroom designs that combine LLM and NMT outputs, automatic metrics, human adequacy/fluency assessment, error annotation, and a consequential post-editing decision within a single assignment. The present study addresses that integrated pedagogical configuration. The importance of combining these components in a single assignment is that the final post-editing choice becomes consequential: students cannot discuss metrics, errors, system affordances, and ethical reliance on AI as separate issues, but must reconcile them in one accountable decision about what to revise and why.

## 3. Methodology

### 3.1. Research design

The study adopts an exploratory qualitative case-study design with descriptive quantitative support, because the aim is not to benchmark systems across a controlled shared test set, but to analyse the reasoning observable in student decisions within an assessed final assignment completed as part of the course. More specifically, the study investigates how students compared multiple system outputs, justified quality judgements, and selected one version for post-editing on the basis of explicit criteria rather than on the basis of automatic metric scores alone. In this respect, the analysis draws on two concepts: evaluative judgement and AI literacy. Evaluative judgement is treated here as students' developing capacity to discern translation quality, justify preferences, and estimate the extent and type of human intervention required. AI literacy is addressed through students' critical engagement with the affordances and limitations of GenAI and NMT systems within a realistic translation workflow.

### 3.2. Course context

The study was conducted in the 4th year undergraduate course Traducció automàtica i postedició at Pompeu Fabra University, within a translation-focused degree programme, during the 2025–2026 academic year. The course drew on Kenny's textbook (Kenny, 2022) and accompanying activities for core MT and post-editing concepts, while newer topics such as large language models were addressed through additional materials and classroom activities. The assignment was situated in the later part of the course, after students had already received instruction on MT, post-editing, and translation quality evaluation during 10 weeks with both theoretical and practical weekly classes. In pedagogical terms, the task functioned as an integrative final assignment: rather than practising isolated techniques, students were asked to combine human translation, system comparison, metric interpretation, human quality assessment, and post-editing in a single workflow. This integrative design is broadly aligned with recent pedagogical proposals that respond to technological change through student-led experimentation, comparative evaluation, and explicit engagement with post-editing and AI literacy (Ayvazyan et al., 2024). The assignment therefore required students to explain why one output, with its particular strengths and weaknesses, was a better post-editing candidate than the alternatives.

### 3.3. Participants and materials

This study draws on 23 student projects produced for the final assignment in the course Traducció automàtica i postedició. All 23 cases are represented in the final evaluative-judgement master table linked below. Twenty-two cases also include a written report that supports qualitative interpretation, while one case lacks the final report and therefore contributes only limited structured information to the descriptive counts. Students worked into Catalan (n = 13) or Spanish (n = 10) and selected short and specialized English Wikipedia texts of approximately 300 words from several domains, most frequently zoology and translation technology. The task was designed to ensure variation in topic while maintaining a broadly comparable level of genre and text type, as students worked with short expository texts from a publicly available encyclopaedic source. The student reports confirm that the assignment was based on the comparison between a self-produced human translation and four machine-generated versions, followed by the analysis of their relative strengths and weaknesses.

The study analyses student-produced coursework from a regular assessed module and reports the findings as an anonymized classroom case study. Formal ethics approval was not obtained before the assignment or the subsequent analysis, and the study therefore does not claim institutional ethics clearance. To minimize risk, all student work was anonymized before research reporting: students are identified only through case codes, no names or demographic identifiers are reported, and no direct quotations from student reports are used. Findings are presented in aggregate form or through paraphrased, non-identifying case summaries. The analysis focuses on patterns of evaluative reasoning in the assignment rather than on the assessment or profiling of individual students.

### 3.4. Assignment procedure

The assignment followed a multi-stage process. First, each student produced a human translation of the selected English source text into Catalan or Spanish. This stage served both as a reference for later evaluation and as a form of pedagogical preparation: by translating the text themselves before consulting AI or MT outputs, students had already made decisions about meaning, syntax, terminology, and style before comparing system outputs. This prior engagement helped them identify the main decision points in the source text and provided a stronger basis for the subsequent evaluation of machine-generated translations. Second, students generated four automatic translations: two with LLM-based systems and two with neural MT systems. In the case of the LLM systems, prompts and system versions were not fully standardized across all cases. While this reflects authentic classroom use of GenAI, it also limits strict technical reproducibility. Across the reports, the exact tools vary somewhat, but the pedagogical structure remained stable: students always compared two systems presented as GenAI and two systems presented as neural MT. Third, the four system outputs were evaluated automatically using the MATEO environment (Vanroy et al., 2023) and the metrics COMET, BLEU, TER, and ChrF2. Fourth, students carried out a human evaluation focused on adequacy and fluency on a scale of 4 at segment level, often accompanied by segment-level comments and error identification. Fifth, students selected the output they considered most suitable for post-editing and justified that choice. Finally, they performed post-editing on that selected version only, typically aiming at a publishable or fully revised target text. Student reports generally confirm this sequence of human translation, four-system comparison, automatic evaluation, human evaluation, system selection, and post-editing.

### 3.5. Data analysed

The analysis draws on several layers of student-produced material. These include the initial human translations, the four AI-generated outputs, the automatic metric scores, the human evaluation scores and comments, the error annotations recorded during evaluation, the post-edited final versions, and the written rationales included in the final reports. In analytical terms, the reports were analysed as evidence of students' reasoning, not just as records of system preference. Particular attention was therefore paid to the criteria students used when justifying preferences, especially where automatic and human evaluations diverged. Across the dataset, students frequently referred to adequacy, fluency, terminology, naturalness, structural calques, and expected post-editing effort when explaining their decisions. For this reason, the reports are central to the analysis: they show how students moved from scores and rankings to justified decisions about quality and revision.

In addition to the student reports themselves, the analysis relied on a final master coding table that consolidated, for each participant, the text domain, target language, systems compared, best automatic-evaluation system, best human-evaluation/final post-editing system, narrative evidence summary, reflection on AI tools, evaluative-judgement coding, and error notes. This master table made it possible to combine descriptive quantitative reporting with cross-case qualitative interpretation.

Descriptive counts based on system frequencies and broad coding distributions include all 23 projects represented in the master table. Interpretive claims about student reasoning, however, are grounded primarily in the 22 cases for which a written report was available.

### 3.6. Analytical procedure

The study was conducted by the course instructor, who designed the assignment and subsequently analysed the student submissions for research purposes. This dual role is methodologically relevant. The study does not claim the neutrality of an external observer; rather, it treats the reports as a pedagogically grounded classroom corpus interpreted reflexively in relation to the course design.

The analysis combined descriptive quantitative aggregation with qualitative cross-case interpretation. Each report was first summarized using a common extraction template covering target language, systems compared, automatic-evaluation winner, student-selected system for post-editing, stated reasons for preference, recurring error types, post-editing effort, and explicit reflection on AI or MT tools. This template made it possible to compare cases systematically while preserving the contextual specificity of each student report.

The coding strategy was hybrid. Deductive categories were derived from the assignment design and the conceptual framing of the study, including adequacy, fluency, metric–human comparison, post-editing effort, and final system selection. Inductive subthemes were then identified through close reading of the reports, especially where students foregrounded issues such as terminology instability, acronym handling, scientific naming, syntactic naturalness, orthotypographic conventions, genre appropriateness, or the limitations of the human reference translation.

The reports were coded at case level rather than at segment level. The first layer of analysis recorded descriptive patterns, including whether the system favoured by automatic metrics was also selected by the student after human evaluation, or whether there was a divergence between metric-based and human judgement. Particular attention was paid to cases in which students explicitly rejected the metrically strongest output in favour of another system, since these moments offered especially clear evidence of evaluative judgement.

A second layer of coding focused on the criteria students used to justify their decisions. These included adequacy, fluency, terminology, naturalness, structural calques, formatting, orthotypographic conventions, grammatical accuracy, omissions, genre fit, and estimated post-editing effort. References to post-editing effort were coded when students commented on the extent of reformulation required, the number or severity of interventions needed, or whether the chosen output allowed for localised rather than extensive revision.

Finally, each case was assigned an evaluative-judgement rating on an ordinal scale: Low, Medium, Medium to High, or High. A report was coded as High when it triangulated several forms of evidence, such as automatic metrics, human adequacy and fluency evaluation, error analysis, and post-editing effort, and used them to justify the final system choice. Medium to High was used when the reasoning was explicit and multidimensional but less fully developed. Medium was assigned when the report included some qualitative justification but remained partly descriptive or metric-led. Low was used when the report mainly reported scores or outcomes without substantial evaluative interpretation.

AI-assisted drafting was used only for linguistic standardization of coding summaries after the researcher had reviewed the reports and assigned the categories; it was not used to assign codes, determine ratings, or generate analytical findings. All analytical decisions, category definitions, evaluative-judgement ratings, and final codes were checked and finalized by the researcher.

## 4. Results

### 4.1. Overview of student submissions

The results are organized in relation to the three research questions: first, the distribution of final selections and selection rationales; second, the relationship between automatic metric rankings and final human/post-editing choices; and third, the forms of evaluative judgement and critical AI use observable in the reports.

The final master table[2] contains 23 student projects, of which 22 are accompanied by written reports substantial enough for qualitative interpretation. The dataset includes 13 Catalan-target and 10 Spanish-target projects. Across the corpus, the most frequently used systems are ChatGPT (22 projects), Gemini (21), and DeepL (20), followed by Softcatalà (11), Bing Translator (8), Google Translate (5), Mistral (3), and eTranslation (2). This confirms that the assignment generated a broad but still comparable set of multi-system comparisons.

[2] Evaluative Judgement Master. The full anonymized coding matrix is available as supplementary material. It includes case identifier, target language, domain, systems compared, automatic-evaluation winner, final selected system, selection rationale, error categories, post-editing effort, reflection on AI tools, and evaluative-judgement level. https://docs.google.com/spreadsheets/d/1BuKgQuMnQ2m8Bqo6XN6dO6KZQsJWtfmS7UckIuzGQ1U/edit?usp=sharing

The distribution of "best" systems differs depending on whether one looks at automatic or human/final selection. In the automatic-evaluation column, ChatGPT is the most frequent best-ranked system (9 cases), followed by DeepL (6) and Gemini (3); there is also one ChatGPT/Gemini tie, plus 2 wins for Bing Translator, 1 for Softcatalà, and 1 for Google Translate. By contrast, in the final human/post-editing selection column, DeepL is chosen most often (10 cases), followed by Gemini (7), ChatGPT (5), and Bing Translator (1). At the family level, final choices are almost evenly split between LLM/GenAI systems (12 cases) and NMT systems (11 cases). These distributions suggest that final preferences were not dominated by a single tool family.

A further descriptive pattern appears at target-language level. In the Catalan-target projects, Gemini is the most frequent final choice (7 of 13), followed by DeepL (3), ChatGPT (2), and Bing Translator (1). In the Spanish-target projects, DeepL is the most frequent final choice (7 of 10), followed by ChatGPT (3). These counts should be interpreted cautiously because the source texts vary across projects, but they still show that student preferences were justified and text-dependent rather than uniform.

### 4.2. Evidence of evaluative judgement

The strongest result in the corpus is not that one system consistently "won," but that the assignment elicited criterion-based evaluative judgement. In the final coding, 14 reports are classified as High in evaluative judgement and 2 as Medium to High; 5 are coded as Medium, and only 1 as Low. The distribution indicates that most reports contained some form of explicit justification, usually involving more than one quality criterion. Table 1 summarizes the main forms of evaluative judgement identified in the report coding.

**Table 1.** Evidence of critical/evaluative judgement across student reports (n = 22 analysable reports)

| Manifestation of evaluative judgement | How it was evidenced in the coding | n/22 | Typical form in the reports | Analytical significance |
|---|---|---|---|---|
| **Evaluative judgement was substantively present in most reports** | Reports coded High or Medium to High in the master table | **16** | Students justified preferences with explicit criteria rather than merely naming a preferred system | Shows that the task elicited genuine judgement, not simple tool preference |
| **Students frequently compared automatic and human evaluation explicitly** | Critical-judgement coding referred to direct comparison between metric results and human assessment | **15** | Students contrasted MATEO rankings with adequacy/fluency judgements and then explained why they converged or diverged | Supports the argument that metrics were used critically rather than accepted uncritically |
| **Students often overrode the automatically best-ranked system** | Best automatic system differed from the final post-editing choice | **11** | A system ranked best automatically was not always selected for post-editing | This is one of the clearest indicators of evaluative judgement in the dataset |
| **Quality was usually treated as multidimensional** | Reports used at least two quality dimensions beyond raw metric ranking | **20** | Students distinguished combinations such as adequacy, fluency, terminology, naturalness, grammar, formatting, coherence, or genre fit | Shows emerging translator reasoning about quality as composite and justified |
| **Post-editing effort frequently entered into the final decision** | Reasons for preference explicitly invoked fewer changes, easier revision, | **14** | Students selected the version that would require fewer or more localized corrections, | Connects judgement to workflow awareness and |
| | lower intervention burden, or faster post-editing | | even when another system scored well automatically | professional usability |
| **A smaller but important group adopted a metaevaluative stance** | Reports explicitly questioned metric interpretation, reference-translation effects, domain limitations, or validated terminology externally | **6** | Students reflected on the limits of TER/COMET, the influence of their own reference translation, or the need to check domain terminology | Marks the strongest form of critical AI literacy in the dataset |

Across the stronger reports, students justified choices by weighing adequacy, fluency, terminology, naturalness, genre fit, semantic precision, and expected post-editing effort against one another. In other words, the assignment made traceable the kind of multidimensional reasoning that professional post-editing requires. This pattern is especially clear in cases where students explicitly rejected the metrically strongest system in favor of the output they judged more suitable for full revision.

At the same time, not all reports showed the same degree of analytical depth. A small minority remained largely metric-descriptive or offered limited qualitative justification. The weaker reports are also informative, because they show where additional scaffolding is needed: especially in connecting metric scores to error analysis and post-editing decisions.

### 4.3. Automatic metrics and human judgement

The clearest quantitative pattern concerns metric/human convergence. In the 22 report-backed cases, the best-ranked system in automatic evaluation and the final selected system matched in 12 cases and diverged in 10. The near-even split shows that metric rankings influenced the reports, but did not determine the final post-editing choice. Instead, students treated the metrics as one source of evidence among others. Table 2 compares the system most strongly favoured by automatic evaluation with the system ultimately selected by students for post-editing, making traceable the extent to which metric-based ranking and final human judgement converged or diverged across submissions. The student who did not submitted the final report is also included in the table. The best automatic system was coded as the system identified as strongest in the student's automatic-evaluation summary. Where students reported conflicting metric rankings, the coding followed COMET priority or the student's own stated automatic winner. Ties were retained where no single automatic winner was identifiable.

**Table 2.** Distribution of automatically best-ranked systems and final human/post-editing selections (n = 23)

| System | Best automatic system (n) | Final selected system (n) | System family |
|---|---|---|---|
| ChatGPT | 9 | 5 | GenAI |
| DeepL | 6 | 10 | NMT |
| Gemini | 3 | 7 | GenAI |
| Bing Translator | 2 | 1 | NMT |
| Google Translate | 1 | 0 | NMT |
| Softcatalà | 1 | 0 | NMT |
| ChatGPT=Gemini (tie) | 1 | 0 | GenAI tie |

This pattern is also visible in the contrast between system frequencies. ChatGPT appears most often as the automatic evaluation winner, but DeepL and Gemini together dominate the final human/post-editing choices. In practice, students often used automatic results as a starting point and then revised their conclusions when human evaluation surfaced issues of terminology, semantic fit, naturalness, or edit burden. Several reports make this tension explicit. In one case (C07), DeepL had stronger BLEU and TER values, but ChatGPT was selected because its Catalan syntax sounded more natural and its scientific terminology required fewer interventions. In another case (C10), ChatGPT was ranked first at the automatic evaluation step, but DeepL was chosen because it avoided a key semantic imprecision and preserved the source structure in a way that would reduce post-editing effort; the same report rejects Gemini partly because its unnecessary restructuring would create extra workload for a real post-editor.

### 4.4. Error annotation findings

The master table also makes it possible to report broader descriptive patterns in the error analyses. A broad consolidation of the errors noticed field shows that terminology issues appear in 19 reports, accuracy in 15 reports, syntactic issues in 14, fluency/style issues in 14, and grammar issues in 13. These groupings are descriptive rather than absolute, but they indicate clearly where students' attention was concentrated. Table 3 summarizes the broad error categories most frequently identified in student reports, based on a cross-case coding of the error descriptions recorded in the master table.

**Table 3.** Broad error categories identified across student reports.

| Error category | Reports (n=22) | Typical problems observed |
|---|---|---|
| Terminology / lexis | 19 | unstable technical terms, lexical inaccuracies, false friends, untranslated items, acronym handling |
| Adequacy / meaning transfer | 15 | omissions, additions, reinterpretation, false sense, conceptual imprecision |
| Syntax / calques / structure | 14 | English-influenced structures, passive voice, literal syntax, awkward sentence structure |
| Naturalness / fluency / style | 14 | less idiomatic phrasing, awkward wording, low stylistic naturalness |
| Grammar / orthography / formatting | 13 | agreement, punctuation, spelling, apostrophation, numeric/notation conventions |

The error analysis shows that students' attention was concentrated not on fluency alone, but on specialized terminology, semantic precision, literal or English-influenced syntax, omissions/additions, and target-language conventions such as orthotypography, numeric formatting, and code or label preservation. One student report (C11), for example, values ChatGPT not only for adequacy and fluency, but also for preserving code, tags, and numbering, while still noting structural calques, anglicisms, and orthotypographic issues. A scientific Catalan report (C22) selects ChatGPT over DeepL because it offers a better overall balance, even though DeepL scores better on BLEU and TER metrics; the student highlights scientific terminology, conceptual precision, notation, and syntactic naturalness as decisive factors. Another report (C21) selects Bing Translator despite the student's initial expectation that the AI systems would perform better, because Bing Translator offered stronger terminological coherence and more natural syntax than the alternatives.

### 4.5. Selection and post-editing effort

The reports show that students usually selected not the output they considered best in the abstract, but the output they considered the best basis for full post-editing. In the final table, the chosen system is very often described as requiring low or low-to-moderate intervention, and students repeatedly explain their decisions in terms of fewer severe errors, less restructuring, better terminology, or more natural target-language syntax.

Many students evaluated outputs as post-editing candidates: the preferred version was often the one that seemed safest and most efficient to revise. That logic is explicit in several reports. One student (C10) prefers DeepL over ChatGPT partly because ChatGPT's errors would require heavier reformulation; the same report rejects Gemini because its restructuring would create unnecessary real-world workload. C09 chooses Gemini even though automatic evaluation leans toward ChatGPT, because Gemini, according to the report, offers the best final quality and needs only a few changes in specialized vocabulary.

### 4.6. AI literacy and critical use

Several reports show conditional use of AI/MT outputs: they accepted useful suggestions, rejected problematic ones, and justified where human revision remained necessary. Where explicit reflection is present, students do not treat AI-generated outputs as inherently superior, nor do they reject them categorically. Instead, they repeatedly evaluate them in relation to text type, terminology, target-language conventions, reference quality, and anticipated post-editing effort. This is consistent with the broader pattern in the master table, where reflections on AI tools tend to frame them as tools that can accelerate parts of the translation process but still require terminological checking, adequacy control, and stylistic revision by a human translator.

A further sign of developing AI literacy is that some students no longer frame the task as choosing a single "perfect" system. Instead, they begin to see different systems as offering partial strengths that can inform a better final result. One report (C08) explicitly comments that each tool contributed something useful to the final translation process, even though one system still had to be selected for full post-editing. This is pedagogically important because it suggests that the student was not choosing a system blindly, but using differences between systems to improve the final translation.

When considered together, these reflections indicate that the assignment appears to have supported, and at least documented, not only system comparison, but also a more critical understanding of when AI output can be useful, where its risks lie, and why human judgement remains necessary. In that sense, the task appears to support AI literacy as a practical form of professional reasoning: students practice deciding when to rely on a system, when to question it, and what kind of human correction is needed.

## 5. Discussion

The results reposition system ranking as a means rather than an end: its main function was to prompt students to justify a post-editing decision. Students' final choices were distributed across both LLM/GenAI and neural MT systems, and the automatically best-ranked system did not always coincide with the version selected for post-editing.

The pattern fits the MT literacy and AI literacy literature, where technological competence is defined as contextual understanding rather than operation alone (O'Brien and Ehrensberger-Dow, 2020; Ehrensberger-Dow et al., 2023; Long and Magerko, 2020). In the present study, students used automatic metrics as one layer of evidence, but stronger reports did not treat metric scores as final judgements. Instead, students weighed them against adequacy, fluency, terminology, target-language naturalness, structural fit, and expected post-editing effort. The task positioned automatic evaluation as an object of interpretation, not an answer key.

The error patterns identified in the reports further confirm the value of combining automatic metrics with human assessment and error annotation. Students repeatedly focused on terminology, semantic precision, calques, passive structures, orthotypographic conventions, numerical formatting, omissions, additions, and target-language idiomaticity. Such issues may be underrepresented in purely score-based evaluation, especially when an output is fluent or close to the reference. In this respect, the assignment builds on earlier pedagogical work on MT evaluation and post-editing (Moorkens, 2018; Öner and Öner Bulut, 2021) by placing similar evaluative practices in a contemporary workflow that includes both LLM and NMT outputs.

The findings also caution against a simplified opposition between LLMs and NMT. The dataset does not support the claim that one system family is categorically superior. Instead, preferences were shaped by target language, domain, terminology, naturalness, and post-editing burden. For teaching, this variability is useful because it requires students to justify preferences in relation to a specific text, language, domain, and revision goal. It encourages students to treat AI and MT systems as resources with different affordances and risks, rather than as stable hierarchies of quality. Such a conclusion is consistent with recent translation-specific discussions of AI literacy, which emphasize suitability, data awareness, prompting, ethics, and contextual decision-making (Bowker, 2026; Inglada, 2026; Yamada, 2026; Moorkens and Doğru, 2026).

Finally, the study suggests one way of operationalizing evaluative judgement, AI literacy, and professional agency in a classroom assignment. Students were required to generate outputs, compare them, interpret metrics, perform human evaluation, identify errors, select one candidate, and post-edit it. This sequence left an analysable record of their reasoning. It also encouraged students to qualify their trust in the systems: outputs were useful, but still had to be checked against meaning, terminology, style, and post-editing effort. In this sense, the assignment operationalizes the idea that students must remain the "arbiters of quality" in AI-mediated translation (Bearman et al., 2024). Its contribution lies in showing that classroom tasks can move students beyond passive tool use toward critical comparison, accountable selection, and professionally grounded revision.

## 6. Conclusion

This case study has shown that the assignment's strongest contribution was to document how students justified translation-quality decisions within a realistic post-editing workflow. Across the 23 projects, preferences were neither uniform nor mechanically determined by automatic metrics. Instead, students justified their choices by weighing adequacy, fluency, terminology, naturalness, structural fit, and anticipated post-editing effort. The most important outcome was the reasoning itself: students connected quality, usability, and revision effort when deciding which output to post-edit.

A second key finding is that automatic metrics functioned as a useful starting point, but not as a sufficient basis for decision-making. In a substantial number of cases, the automatically best-ranked system was not the one ultimately selected for post-editing, because human evaluation brought into view issues that metrics alone could not capture adequately, especially semantic precision, terminology, target-language naturalness, restructuring, and local conventions. In this sense, the assignment helped students move beyond score comparison toward a more professionally relevant understanding of quality as multidimensional, contextual, and closely tied to revision effort.

The findings therefore support the use of multi-system comparison as a pedagogical design for AI literacy in translator education. The task encouraged a more conditional stance toward AI: students learned to use machine output when it was useful, question it when it was misleading, and justify the need for human intervention. In doing so, they learned to question fluent-looking output, interpret metrics critically, distinguish understandable from professionally acceptable translation, and justify why one version was preferable for a specific text and purpose.

This paper contributes: (1) an adaptable classroom workflow for comparing LLM and NMT outputs through automatic and human evaluation; (2) descriptive and qualitative evidence of how students justify post-editing selections; and (3) an argument for treating evaluative judgement as a practical component of AI literacy in translator education.

These conclusions should be read in light of the study's limits. The dataset is small, classroom-based, and built on text-specific comparisons using student-produced reference translations, so the paper does not support broad claims about whether LLMs or NMT systems are categorically superior. Its contribution is pedagogical: it shows how a comparison-and-post-editing task can generate evidence of students' evaluative reasoning. In this setting, students become less passive users of AI-based translation tools and more critical comparators, revisers, and arbiters of quality.